\documentclass[10pt,twocolumn,letterpaper]{article}
\usepackage{iccv,times,epsfig,graphicx,amsmath,amssymb,algorithm2e,rotating,caption,subcaption,booktabs,multirow,floatrow,makecell}
\usepackage[pagebackref=true,breaklinks=true,colorlinks,bookmarks=false]{hyperref}
\newcommand{\plottsne}[2]{
  \begin{subfigure}{0.425\textwidth}
    \includegraphics[width=\textwidth]{figures/#1.pdf}
    \vspace{-5mm}\caption{#2}\vspace{1mm}
  \end{subfigure}}
\newlength\savewidth\newcommand\shline{\noalign{\global\savewidth\arrayrulewidth
  \global\arrayrulewidth 1pt}\hline\noalign{\global\arrayrulewidth\savewidth}}

\iccvfinalcopy
\def\iccvPaperID{6367}

\ificcvfinal\fi

\begin{document}
%%%%%%%%%%%%%%%%%%%%%%%%%%%%%%%%%%%%%%%%%%%%%%%%%%%%%%%%%%%%%%%%%%%%%%%%%%%%%%%
% TITLE
%%%%%%%%%%%%%%%%%%%%%%%%%%%%%%%%%%%%%%%%%%%%%%%%%%%%%%%%%%%%%%%%%%%%%%%%%%%%%%%
\title{Domain-Adaptive Single-View 3D Reconstruction}
\author{Pedro O. Pinheiro\\
Element AI\\
\and
Negar Rostamzadeh\\
Element AI\\
\and
Sungjin Ahn\\
Rutgers University\\
}

\maketitle
\ificcvfinal\thispagestyle{empty}\fi

%%%%%%%%%%%%%%%%%%%%%%%%%%%%%%%%%%%%%%%%%%%%%%%%%%%%%%%%%%%%%%%%%%%%%%%%%%%%%%%
% ABSTRACT
%%%%%%%%%%%%%%%%%%%%%%%%%%%%%%%%%%%%%%%%%%%%%%%%%%%%%%%%%%%%%%%%%%%%%%%%%%%%%%%
\begin{abstract}
Single-view 3D shape reconstruction is an important but challenging problem, mainly for two reasons. First, as shape annotation is very expensive to acquire, current methods rely on synthetic data, in which ground-truth 3D annotation is easy to obtain. However, this results in domain adaptation problem when applied to natural images. The second challenge is that there are multiple shapes that can explain a given 2D image. In this paper, we propose a framework to improve over these challenges using adversarial training. On one hand, we impose domain confusion between natural and synthetic image representations to reduce the distribution gap. On the other hand, we impose the reconstruction to be `realistic' by forcing it to lie on a (learned) manifold of realistic object shapes. Our experiments show that these constraints improve performance by a large margin over baseline reconstruction models. We achieve results competitive with the state of the art with a much simpler architecture.
\end{abstract}

%%%%%%%%%%%%%%%%%%%%%%%%%%%%%%%%%%%%%%%%%%%%%%%%%%%%%%%%%%%%%%%%%%%%%%%%%%%%%%%
% Introduction
%%%%%%%%%%%%%%%%%%%%%%%%%%%%%%%%%%%%%%%%%%%%%%%%%%%%%%%%%%%%%%%%%%%%%%%%%%%%%%%
\section{Introduction}\label{sec:introduction}
% Humans do it -- rely on shape priors
Humans can easily understand the underlying 3D structure of scenes and objects from single images. This is a hallmark of a human visual system and it is an essential step towards higher level visual understanding.
This is an extremely ill-posed problem because a single image does not contain enough information to allow 3D reconstruction. Therefore, a machine vision system needs to rely on priors over the shape to infer 3D structure.

Efficient and effective 3D prototyping plays an important role in many different fields, such as virtual/augmented reality, architecture, robotics and 3D printing to name a few. Perhaps more importantly, studying 3D object representations could bring insights on how this information is encoded in intermediate and higher-level visual cortices~\cite{yukako08shape,kourtzi11neural}.

% old school reconstruction
Traditional reconstruction methods rely on multiple images of same object instance~\cite{laurentini1994hull,bonetV99,broadhurst2001APF,seitz2006,furukawa2015multiview}. These methods possess two strong limitations due to some key assumptions~\cite{choy2016r2n2}: (i) it requires a large number of views to achieve reconstruction, (ii) the objects' appearance are expected to be Lambertian (\ie, non-reflective) and their albedos are supposed to be non-uniform (\ie, rich of non-homogeneous textures).

% Use of priors (shape prior)
Another way to achieve 3D reconstruction is to leverage knowledge from object's appearance and shape. The main advantages of relying on \emph{shape priors} is that we do not need to rely on accurate feature correspondences across different views. In this case 3D reconstruction can, in principle, be done from a single-view 2D image (assuming the priors are rich enough).
% maybe mention some classical methods from r2n2 paper?

\begin{figure}[!t]
\begin{center}
      \includegraphics[width=1\linewidth]{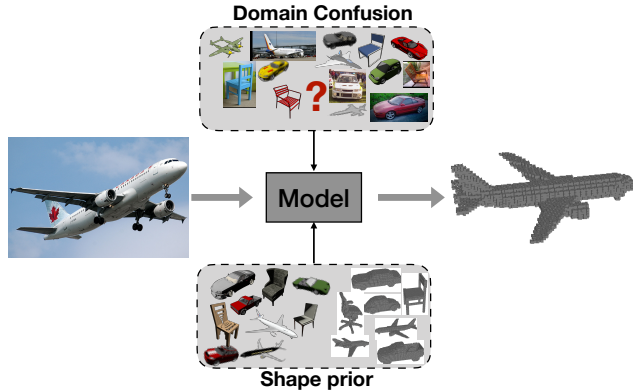}
      % \rule{6cm}{6cm}
\end{center}
   \caption{We propose a framework for (natural) single-view 3D reconstruction exploiting adversarial training in two ways. These constraints are achieved with additional loss terms. We impose domain confusion between natural and rendered images (top) and exploit shape priors to force reconstructions to look realistic (bottom).}
\label{fig:summary}
\end{figure}

% large dataset + deep learning make it eexciting
Recently, there has been a growing interest in learning-based approaches to tackle the problem of predicting the canonical shape of an object from a single image~\cite{kar2015csrec,choy2016r2n2,girdhar16tnet,tatarchenko2016multiview,xinchen2016perspective,
rezende2016structure3d,wu2016gan3d,novotny2017look,tulsiani2017drc,wu2017marrnet,wu2018shapeprior,
yang2018posesup}.
Two technical advances were responsible for this surge: (i) the easy access to large-scale 3D Computer-Aided Design (CAD) repositories, such as ShapeNet~\cite{chang2015shapenet}, Pascal3D+~\cite{xiang14pascal3d}, ObjectNet3D~\cite{xiang2016objectnet3d}, Pix3D~\cite{sun2018pix3d} and (ii) advances in deep learning techniques~\cite{goodfellow16dl}.

% current methods
Most of these methods contain a similar high-level architecture that regresses a 3D shape from (rendered) images: an encoder transforms a 2D image into a latent representation and a decoder reconstructs the 3D representation. They differentiate in how constraints from 3D world are imposed, \eg,~\cite{choy2016r2n2,xinchen2016perspective,tulsiani2017drc} force multi-view consistency to learn the 3D representation, while~\cite{wu2017marrnet,wu2018shapeprior} make use of 2.5D sketches.
These approaches use a large number of CAD models to leverage shape priors (either making explicit use of 3D representation or not).

% and their issues (due to domain adaptation)
Single-view 3D reconstruction is a very ill-posed problem.
In order to learn strong shape priors to infer 3D structure, deep learning methods require a large amount of 3D object annotations.
However, acquiring good 3D object annotation from natural images is an extremely challenging endeavor. Most deep learning approaches, therefore, make use of synthetic images (which can be rendered easily if a proper 3D representation is given).

Convolutional neural networks (CNNs)~\cite{lecun98cnn} are known to perform sub-optimally when the data distribution of inputs changes, a problem known in the computer vision literature as \emph{domain shift}~\cite{torralba11cvpr}.
For this reason, CNN-based 3D reconstruction, trained on synthetic images, performs worse when applied to natural images.

% what we propose
In this paper, we introduce a method to improve the performance of reconstruction models in natural images, where proper 3D labels are very difficult to acquire. To achieve this goal, we impose two constraints on the network's reconstruction loss (expressed as additional loss terms) based on shape prior learned from large 3D CAD repository (see Figure~\ref{fig:summary}).

First, inspired by the domain adaptation literature~\cite{csurka17uda,ganin16revgrad}, we force the encoded 2D features to be invariant with respect to the domain they come from (rendered or natural). This way, a decoder trained on synthetic images will naturally perform better on real images. Second, we constraint the encoded 2D features to lie in the manifold of realistic objects shapes. This constraint forces the decoded 3D reconstruction to look more realistic. These two loss terms are characterized through adversarial training~\cite{goodfellow2014gan,ganin16revgrad}, an active research topic.

% contributions
Our main contributions can be summarized as follows: (i) we propose a model and a loss function that exploit learned shape priors to improve performance of natural image 3D reconstructions (using adversarial training in two different ways), (ii) we show that this method boost performance in both voxel and point cloud representations, and (iii) the proposed method achieves results competitive with state of the art on different datasets, with a much simpler architecture.
Moreover, the proposed approach is independent of the encoder-decoder architecture and can be applied to different single-view 3D reconstruction models.

% rest of the paper
The rest of the paper is organized as follows: Section~\ref{sec:related-work} presents related work, Section~\ref{sec:method} describes how we learn the shape prior and leverage it in two different ways for learning reconstruction, and Section~\ref{sec:experiments} describes our experiments in different datasets. We conclude in Section~\ref{sec:conclusion}.

%%%%%%%%%%%%%%%%%%%%%%%%%%%%%%%%%%%%%%%%%%%%%%%%%%%%%%%%%%%%%%%%%%%%%%%%%%%%%%%
% Related Work
%%%%%%%%%%%%%%%%%%%%%%%%%%%%%%%%%%%%%%%%%%%%%%%%%%%%%%%%%%%%%%%%%%%%%%%%%%%%%%%
\section{Related Work}\label{sec:related-work}

\paragraph{Single-view 3D reconstruction.}
Traditional reconstruction methods rely on multiple images of same object instance to achieve reconstruction~\cite{laurentini1994hull,bonetV99,broadhurst2001APF,seitz2006,furukawa2015multiview}. Recently, data-driven approaches to 3D reconstruction from single image have appeared.
These methods can roughly be divided into two types: (i) those that explicitly use 3D structures~\cite{girdhar16tnet,choy2016r2n2,wu2016gan3d,fan2017points,groueix2018atlas,wu2017marrnet,wu2015objectnet} and (ii) those that use other sources of information to infer the 3D structure~\cite{vicente14recvoc,kar2015csrec,xinchen2016perspective,rezende2016structure3d,gwak17weakly,broadhurst2001APF,tulsiani2017drc,yang2018posesup}.

These approaches, based on deep learning techniques, usually share a similar (high-level) architecture: an encoder that maps 2D (rendered) images into a latent representation and a decoder that maps this representation into a 3D object. They tend to differ in the way 3D world constraints are imposed.
For instance,~\cite{choy2016r2n2,xinchen2016perspective,xinchen2016perspective, tulsiani2017drc,gwak17weakly,rezende2016structure3d,kundu18rcnn3d} force multiview consistency to learn the 3D representation, while~\cite{vicente14recvoc,kar2015csrec,kanazawaT18collections} leverage keypoints and silhouette annotations. Other approaches~\cite{wu2017marrnet,wu2018shapeprior} leverage 2.5D sketches (surface normals, depth and silhouette) information to improve prediction.

More recently, Zhang, Zhang \emph{et. al.}~\cite{zhang18unseen} consider spherical maps (in additional to 2.5D sketches) to learn 3D representations.
Contrary to most work on single-view 3D reconstruction, the proposed method does not use canonical shape: every ground-truth 3D representation is on the same viewpoint as the 2D training sample.
This work is the first to look at reconstructing shapes for unseen classes, however, it does not deal with domain-adaptation issues.

Contrary to all these methods, our approach does not use any additional information besides RGB images. However, in addition to rendered images, we also use  unlabeled natural images (which are easy to acquire). We note that our contributions are independent of the encoder and decoder architecture (as long as they are differentiable), and could be applied in many of these more powerful encoder-decoder architectures. In experiments, we show that our approach improves performance over two baselines: a simple voxel encoder-decoder architecture and AtlasNet~\cite{groueix2018atlas}, a state-of-the-art encoder-decoder architecture based on point clouds representation.

\paragraph{Domain adaptation.}
The difficulty to acquire 3D annotations for natural images forces reconstruction models to learn from rendered images. It is well known in the literature~\cite{torralba11cvpr,csurka17uda} that the performance of a model drops if applied in data coming from a distribution different from the one used during training.
Ganin \emph{et al.}~\cite{ganin16revgrad} deal with this issue by forcing domain confusion (between two domains) through an adversarial objective. Many works have been dealing with domain adaptation from synthetic to real for image classification~\cite{saito18advdrop,sait18maxdisc,pinheiro18simnet,saito18openset}.

In this work, we borrow ideas from domain adaptation literature to impose domain confusion in a similar way as these previous work. We consider, however, the more challenging problem of 3D reconstruction instead of simple image classification.

\paragraph{Shape priors.}
Reconstruction of 3D structure from single-view images requires strong priors about object's shape. Many works focus on better capturing the manifold of realistic shapes. Non-deep approaches had focus on low-dimensional parametric models~\cite{blanz99morphable,kar2015csrec}.The authors of~\cite{girdhar16tnet,li15joint} use CNNs to learn a common embedding space for 2D rendered images and 3D shapes. Other methods rely on generative modeling to learn shape prior, \eg,~\cite{wu2015objectnet} use deep belief nets to model 3D representations, ~\cite{rezende2016structure3d, broadhurst2001APF, eslami18gqn} consider variants of variational autoencoders and~\cite{wu2016gan3d} use a variant of GANs~\cite{goodfellow2014gan} to capture the manifold of shapes. In~\cite{makhzani2016aae}, the authors propose an adversarial autoencoder that uses adversarial training techniques to match aggregated posterior to perform variational inference.

A few works use adversarial training for single-view 3D reconstruction. Gwak \emph{et al.}~\cite{gwak17weakly} use GANs to model 2D projections instead of 3D shapes.
More similar to our work, Wu, Zhang~\emph{et al.}~\cite{wu2018shapeprior} use adversarial training techniques to impose reconstructions to look more natural. They use  the discriminator of a pre-trained 3D GAN~\cite{wu2016gan3d} to determine whether a shape is realistic.
This approach is similar in principle to one of our contributions. It is, however, implemented in very different way. The input to the discriminator is a high dimensional 3D shape, which makes the training to be very unstable. In our method, the input is a single vector in a low-dimensional space.

%%%%%%%%%%%%%%%%%%%%%%%%%%%%%%%%%%%%%%%%%%%%%%%%%%%%%%%%%%%%%%%%%%%%%%%%%%%%%%%
% Method
%%%%%%%%%%%%%%%%%%%%%%%%%%%%%%%%%%%%%%%%%%%%%%%%%%%%%%%%%%%%%%%%%%%%%%%%%%%%%%%
\section{Method}\label{sec:method}
In our reconstruction setting, we are interested in predicting a \emph{volumetric representation} $v^n\in\mathcal{V}$ from a canonical view of a natural image $x^n\in\mathcal{I}^n\subset \mathbb{R}^{3\times H \times W}$.
In our experiments, the volumetric representation is either voxel ($\mathcal{V}\subset\{0,1\}^{d_{v} \times d_{v} \times d_{v}}$) or point cloud ($\mathcal{V}\subset\mathbb{R}^{d_v\times 3}$).

At training time, we have access to a large repository of 3D CAD objects, where pairs of rendered images and volumetric representation $\mathcal{D}_{rend}=\{(x_i^r,v_i)\}_{i=1}^{N_r}$ are drawn from a distribution $p_r(x,v)$, and unlabeled natural images, $\mathcal{D}_{nat}=\{x_j^n\}_{j=1}^{N_n}$, from a different distribution $p_n(x,v)$.
We note that during training the model has access to natural images (which are easy to acquire), but not their voxel occupancy grid (which are very difficult to gather).

The proposed method, dubbed \emph{Domain-Adaptive REConstruction network} (DAREC), is composed of two components: (i) a \emph{shape autoencoder}, responsible for learning a rich latent representation of 3D objects and (ii) a \emph{reconstruction network}, responsible for inferring the voxel occupancy grid from a 2D image.

% The shape autoencoder
The shape autoencoder is made of an encoder $E$ and a decoder $D$.
The encoder maps 3D representation $v\in\mathcal{V}$ into a low-dimensional embedding representation $e\in\mathcal{E}\subset \mathbb{R}^{d_{e}}$. The decoder maps a data point in the latent space back to a 3D representation.
The voxel shape autoencoder is trained by minimizing the $L_2$ reconstruction loss. The point cloud shape autoencoder is trained by minimizing the Chamfer distance between predicted and ground truth points.

Since the shape autoencoder is trained with true 3D shapes, the learned latent representation lies in the \emph{shape manifold} $\mathcal{E}$, containing low-dimensional embeddings of `realistic' shapes. This component is trained prior to the training of the reconstruction network. The shape prior information is implicitly encoded in this rich representation space.

% reconstruction encoder
The reconstruction network also possesses an encoder-decoder architecture. The encoder $f$, parameterized by $\theta_f$, is responsible to transform a 2D image into an embedding space from which a 3D representation can be reconstructed with a decoder. At inference time, the reconstruction network is the sole network used to predict the voxel occupancy of a given natural test image.

\begin{figure*}[!t]
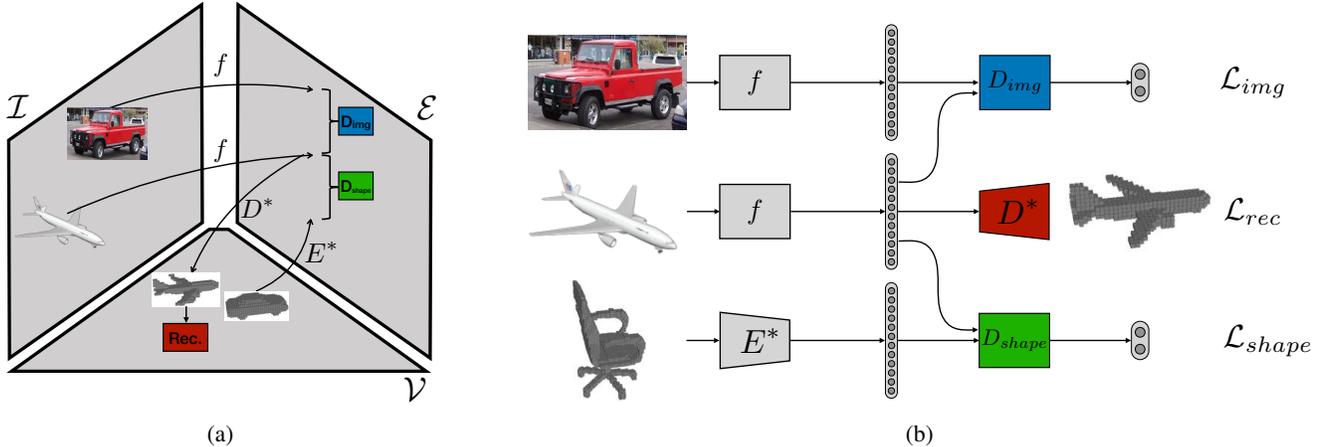

   \centering
   \begin{subfigure}[t]{0.33\textwidth}
      \centering
       \includegraphics[width=\textwidth]{figures/mappings-compressed.pdf}
       \caption{}
       \label{fig:mapping}
   \end{subfigure}%
   ~\hspace{1cm}
   \begin{subfigure}[t]{0.6\textwidth}
      \centering
       \includegraphics[width=\textwidth]{figures/model-compressed.pdf}
       \caption{}
       \label{fig:model_arch}
   \end{subfigure}
\caption{(a) The reconstruction network maps an image to a rich embedding space $\mathcal{E}$ which is then decoded into a 3D shape with the shape decoder $D^*$ (the star indicates we do not update the parameters of $D$ and $E$). The two constraints are imposed on the embedding space with the help of two discriminators: $D_{img}$ imposes image domain confusion and $D_{shape}$ forces the embeddings to lie on the shape manifold. (b) Overview of the proposed architecture.}
\label{fig:model}
\end{figure*}

% General idea of loss
The model is trained in a way that the encoder mapping $f:\mathcal{I}\to\mathcal{E}$ can, at the same time: (i) reconstruct a 3D  representation given a rendered image, (ii) be indistinguishable w.r.t. the domain that the image comes from (either synthetic or real) and (iii) stay in the manifold of `realistic' shapes (learned with the shape autoencoder). To impose these constraints, we define and add the relevant terms to the loss function. Figure~\ref{fig:model} shows an overview of the approach.

% reconstruction loss
The reconstruction loss, $\mathcal{L}_{rec}$, is applied to tuples of rendered images and 3D representations (from $\mathcal{D}_{rend}$).
We use the $L_2$ for reconstruction loss when considering voxel representation and the Chamfer distance (as in~\cite{fan2017points,groueix2018atlas}) for the point cloud representation.
We opt to \emph{not} update the decoder parameters at this training stage. This design choice, combined with the constraint imposed by the third loss, forces the image representations to lie on the manifold of `realistic' shapes.
% \pedro{add equations for rec. vox and pc here?}

In the rest of this section, we show how we leverage adversarial training techniques and (learned) shape prior to improve performance of natural image 3D reconstruction.

%%%%%%%%%%%%%%%%%%%%%%%%%%%%%%%%%%
% Confusing Image Domains
%%%%%%%%%%%%%%%%%%%%%%%%%%%%%%%%%%
\subsection{Confusing Image Domains}
It is well known that machine learning algorithms suffer from domain shift~\cite{torralba11cvpr}. Therefore, a model trained to reconstruct 3D shape from rendered images performs sub-optimally when applied to natural ones.

Theoretical studies~\cite{bendavid2007da,bendavid2010da} suggest that a good cross-domain representation is one in which input domain cannot be easily identified.
We implement such domain confusion by mapping cross-domain features into a common space through adversarial training.
We cast this problem as a minimax game between a domain classifier and feature encoders. That is, we encourage the feature encoder to learn features $f(x)$ that maximize the domain confusion between natural and rendered images.

We consider a discriminator $D_{img}$, parameterized by $\theta_{img}$. The discriminator that classifies the domain of a given feature vector, is optimized by the  standard adversarial classification loss as follows:
\begin{equation}
\begin{split}
\mathcal{L}_{img}(\theta_f,\theta_{img})&=
-\underset{x^{r}\sim p_r}{\mathbb{E}} \text{log}\;D_{img}(f(x^r))\;+ \\
&\;\;\;\;
-\underset{x^{n}\sim p_n}{\mathbb{E}} \text{log}\;(1-D_{img}(f(x^n)))\;.
\end{split}
\label{eq:loss_disc_img}
\end{equation}

We achieve domain confusion by applying Reverse Gradient algorithm~\cite{ganin16revgrad}, which optimizes the parameters $\theta_f$ to maximize the discriminator loss directly, while $\theta_{img}$ minimizes it.

%%%%%%%%%%%%%%%%%%%%%%%%%%%%%%%%%%
% Exploiting Shape Priors
%%%%%%%%%%%%%%%%%%%%%%%%%%%%%%%%%%
\subsection{Exploiting Shape Priors}
A lot of inherent ambiguity exists in single image reconstruction. Multiple objects exists that can explain a single view. For this reason, as noted by Wu, Zhang~\emph{et al.}~\cite{wu2018shapeprior}, 3D reconstruction with only supervised loss tends to predict unrealistic mean shapes.

We characterized a representation to be `realistic' if it belongs close to the manifold created by the (learned) shape autoencoder. We argue that if the feature of a single 2D image $f(x)$ lies in the same manifold, a realistic reconstruction can be achieved by leveraging the decoder of the shape autoencoder.

The third component of our loss, $\mathcal{L}_{shape}$, imposes this constraint by penalizing the model if the distribution of latent embeddings does not match that of the points in the shape manifold. We rely on the learned shape autoencoder to sample these points. Again, we use adversarial training to optimize the loss.

Similar to Equation~\ref{eq:loss_disc_img}, we train a discriminator $D_{shape}$ (parameterized by $\theta_{shape}$) to classify whether a sample is drawn from a 2D encoding representation or from the shape manifold.
Samples from the shape manifold are generated by sampling voxel (or point cloud) instances from ShapeNet and mapping them to the $\mathcal{E}$, using the learned shape encoder $E^*$. The star means that the parameters of the encoder are kept unchanged during this stage of training. This way, we guarantee the encoded samples lie on the learned manifold.

Learning is achieved by minimizing the following loss:
\begin{equation}
\begin{split}
\mathcal{L}_{shape}(\theta_f,\theta_{shape})&=
-\underset{x^r\sim p_r}{\mathbb{E}} \text{log}\;D_{shape}(f(x^r))\;+ \\
&\;\;\;\;
-\underset{v\sim p_r}{\mathbb{E}} \text{log}\;(1-D_{shape}(E^*(v^r)))\;.
\end{split}
\end{equation}

As before, the parameters $\theta_{shape}$ are optimized to minimize this loss while the parameters $\theta_f$ maximize it, therefore, forcing the 2D embeddings to lie on the shape manifold.

%%%%%%%%%%%%%%%%%%%%%%%%%%%%%%%%%%
% Training Details
%%%%%%%%%%%%%%%%%%%%%%%%%%%%%%%%%%
\subsection{Training Details}
The training procedure is done in two stages.

% training the shape autoencoder
We start by training the shape autoencoder to learn shape priors. As we want to capture the intrinsic shape complexity of different objects, we train the model using the full ShapeNet dataset.
We use a different shape autoencoder for each 3D representation considered.

The voxel autoencoder has an encoder $E$ composed of four 3D convolutional layers, each followed by a max-pooling and ReLU~\cite{nair10relu} non-linearity. The first layer contains $5\times5$ filters while the remaining have $3\times3$. The number of hidden units are 32, 64, 128 and 256 respectively.
Similarly, the voxel decoder $D$ has four convolution layers, but instead of max-pooling, we use bilinear upsampling. The dimension of the latent representation is 256.

We use AtlasNet~\cite{groueix2018atlas}~\footnote{we use the official code provided at \url{https://github.com/ThibaultGROUEIX/AtlasNet}} for the point cloud autoencoder. The encoder, similar to PointNet~\cite{qi2017pointnet}, transforms the input point cloud into a latent representation of dimension 1024. The decoder contains four fully-connected layers of size 1024, 512, 256, 128 with ReLU non-linearities (except the last layer, which has a tanh).

Once training converges, we freeze the parameters of the encoder and the decoder and use them in the reconstruction step.

% training the reconstruction network
The architecture of the reconstruction network is shown on Figure~\ref{fig:model_arch}. The parameters of network $f$ are initialized with a ResNet-50~\cite{he16resnet} that was pre-trained to perform classification on ImageNet dataset~\cite{deng09imagenet}. We replace the classification layer by a randomly initialized layer that outputs a vector with dimension of the latent space.

The two discriminators $D_{img}$ and $D_{shape}$ map the embedded features to the probability of which domain the input comes from (modeled by a softmax~\cite{bridle90softmax}). We use two fully-connected layers of dimension 1024, followed by ReLU. We choose not to share but have different set of parameters between the two discriminators because it performs sightly better in practice.

% final objective
Finally the model is optimized to learn 3D representations that are domain-invariant and that lie in the manifold from the prior of realistic shapes. Consequently, our final goal is to optimize the following objective:
\begin{equation}
\begin{split}
\min_{\theta_f} \; \max_{\theta_{img},\theta_{shape}} &
\;\;\mathcal{L}_{rec}(\theta_f) \\
& - \lambda_i\mathcal{L}_{img}(\theta_f,\theta_{img}) \\[2mm]
& - \lambda_s\mathcal{L}_{shape}(\theta_f,\theta_{shape}) \;,
\end{split}
\end{equation}
where $\lambda_{i}$ and $\lambda_{s}$ are balance parameters between the loss terms. We chose $\lambda_{i}$ and $\lambda_{s}$ to be both 0.001 when considering voxel representation and 0.01 with point cloud representation.
To optimize, we used Adam~\cite{kingma14adam} with learning rate of $10^{-4}$ for voxel and $10^{-5}$ for point cloud representation.

%%%%%%%%%%%%%%%%%%%%%%%%%%%%%%%%%%%%%%%%%%%%%%%%%%%%%%%%%%%%%%%%%%%%%%%%%%%%%%%
% Experiments
%%%%%%%%%%%%%%%%%%%%%%%%%%%%%%%%%%%%%%%%%%%%%%%%%%%%%%%%%%%%%%%%%%%%%%%%%%%%%%%
\section{Experiments}\label{sec:experiments}
In this section, we start by comparing the performance of our approach with other methods on the problem of single-view reconstruction from natural images.
We report results with two variants of the model: DAREC-vox, which predicts voxel representations and DAREC-pc, which predicts point cloud representations.
We evaluate the models in two important datasets: the recently released Pix3D~\cite{sun2018pix3d} and PASCAL 3D+~\cite{xiang14pascal3d}.
Then, we study how DAREC behaves with respect to the different loss terms.
Finally, we analyze the learned representation and show qualitative results that corroborates with the notion of domain confusion and shape manifold.

%%%%%%%%%%%%%%%%%%%%%%%%%%%%%%%%%%
% Experimental Setup
%%%%%%%%%%%%%%%%%%%%%%%%%%%%%%%%%%
\subsection{Experimental Setup}
\paragraph{Voxels and point clouds.}
In the first stage of training, we learn shape priors by training a shape autoencoder for the two 3D representations considered.
In both cases, the autoencoder is trained to reconstruct the shape (voxels or point clouds) from ShapeNet dataset~\cite{chang2015shapenet} (we use the ShapeNetCore subset). This dataset contains over 50k object instances of 55 categories.
We use a voxel resolution of $32^3$ (a downsampled version of the voxels provided by the official repository) and 2500 points in the point cloud representation.

The second training stage is responsible for inferring shape representation from a  single-view image. We train two versions of our model: (i) DAREC-vox, which outputs voxel representations and uses the voxel autoencoder and (ii) DAREC-pc, which regresses point clouds and uses AtlasNet for the point cloud shape autoencoder.
In this step, we make use of both natural and rendered 2D images.
We follow previous work and use the same rendered view provided by~\cite{choy2016r2n2}. This allow a more fair comparison between the proposed method and other approaches.
Since we evaluate the model in natural images, we use all rendered data for training.
\vspace{-5mm}
\paragraph{Evaluation metrics.} We evaluate the performance of our method using two metrics: Intersection over Union (IoU) and Chamfer Distance (CD). The metric IoU measures the similarity between ground-truth and (discretized) reconstruction voxels. This is the 3D extension of the common metric (of same name) used in segmentation. The Chamfer distance between two point clouds $P_1,P_2\subset\mathbb{R}^3$ is defined as:
\begin{equation}
\resizebox{1\columnwidth}{!}{
$CD(P_1,P_2) =
\frac{1}{|P_1|}\underset{x\in P_1}{\sum}\underset{y\in P_2}{\min} ||x-y||_2
+\frac{1}{|P_2|}\underset{x\in P_2}{\sum}\underset{y\in P_1}{\min} ||x-y||_2.$
}
\end{equation}

For each point in each set, CD finds the closest point (in the other set) and average the distances. When dealing with voxel occupancy, we first sample points in the the voxel isosurface before computing CD.
It is shown by Sun, Wu \emph{et al.}~\cite{sun2018pix3d} that CD better correlates with human perception. For fair comparison, in the following sections we use the same evaluation code provided by the authors of Pix3D\footnote{\url{http://pix3d.csail.mit.edu/}}.
% \pedro{mention the value of the threshold to binarize the output}

%%%%%%%%%%%%%%%%%%%%%%%%%%%%%%%%%%
% Comparison to Other Methods
%%%%%%%%%%%%%%%%%%%%%%%%%%%%%%%%%%
\subsection{Comparison to Other Methods}
\paragraph{Reconstruction on Pix3D.}
Pix3D is a large-scale benchmark of diverse image-shape pairs with pixel-level 2D-3D alignment.~A significant part of the dataset is chairs because they are common and highly diverse. Following the previous works~\cite{sun2018pix3d,wu2018shapeprior}, we evaluate our approach on the 2,894 untruncated and unoccluded `chair' images.

\begin{figure}[!t]
\begin{center}
    \includegraphics[width=\linewidth]{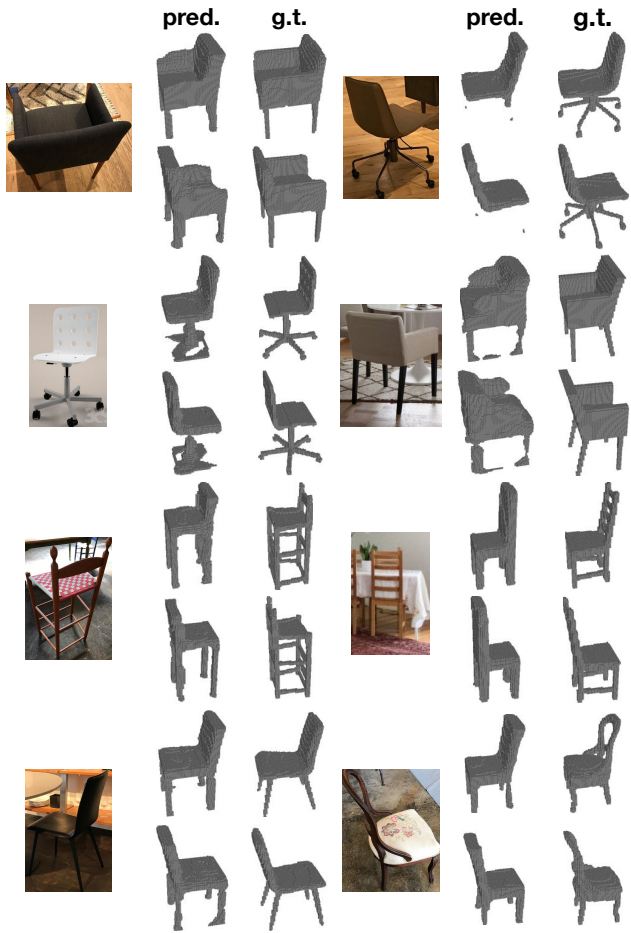}
\end{center}
   \caption{3D reconstruction from single image on Pix3D dataset. For each image, we show the predicted and the ground-truth voxel representations. Our method is capable of learning shape with very different appearances. We show two different views for each 3D representation.}
\label{fig:pix3d-qualitative}
\end{figure}

During training, the reconstruction network has access to synthetic ShapeNet renderings (and their corresponding ground-truth, voxels or point clouds) and unlabeled natural images of `chair' category (we use the natural images of the PASCAL 3D+ r1.1, which contains also ImageNet images). Figure~\ref{fig:pix3d-qualitative} shows the qualitative results of voxel reconstructions generated by our approach. As illustrated in this figure, DAREC is able to reconstruct even in situations of strong self-occlusion.

Table~\ref{table:pix3d} compares the performance of our approach with different methods on the Pix3D dataset. We show results on both IoU (higher is better) and CD (lower is better) metrics. Results from other models are taken from Wu, Zhang \emph{et al.}~\cite{wu2018shapeprior}.

It is also important to mention that these methods use different types of data during training. For instance, PSGN~\cite{fan2017points} require ground-truth masks as input. MarrNet~\cite{wu2017marrnet}, DRC~\cite{tulsiani2017drc} and ShapeHD~\cite{wu2018shapeprior} use depth, surface normals and silhouettes during training.
DAREC achieves competitive results using only RGB images as input and with a much simpler architecture.

\begin{table}[t]
\centering
\begin{tabular}{l|cc}
& IoU & CD \\
\shline
3D-R2N2\cite{choy2016r2n2} &  0.136 & 0.239 \\
3D-VAE-GAN~\cite{wu2016gan3d} &  0.171 & 0.182 \\
PSGN$^{*}$~\cite{fan2017points} &  - & 0.199 \\
MarrNet$^{\dagger}$~\cite{wu2017marrnet} &  0.231 & 0.144 \\
DRC$^{\dagger}$~\cite{tulsiani2017drc} &  0.265 & 0.160 \\
AtlasNet~\cite{groueix2018atlas} &  - & 0.148 \\
AtlasNet + g.t. mask$^{*}$~\cite{groueix2018atlas} &  - & 0.126 \\
ShapeHD$^{\dagger}$~\cite{wu2018shapeprior} &  0.284 & 0.123 \\
\hline
DAREC-vox &  0.241 & 0.140\\
DAREC-pc &  - & 0.112\\
\end{tabular}
\caption{Single-view 3D reconstruction results on Pix3D. We show results on both IoU and CD metrics. $^{*}$~PSGN require ground-truth mask as input. $^{\dagger}$~MarrNet, DRC and ShapeHD use 2.5D sketches to guide training.  Our approach only considers (easily available) natural images during training. We show competitive results in both metrics.}
\label{table:pix3d}
\end{table}

\paragraph{Reconstruction on Pascal 3D+.}
PASCAL 3D+~\cite{xiang14pascal3d} provides annotations for (rough) 3D shape of different rigid object instances from PASCAL VOC 2012~\cite{everingham2015pascal}. Each category has a small set of about 10 CADs per category.

Similar to most of the recent works, we do not use any of the PASCAL 3D+ training set. We use the CAD annotations only for benchmarking purposes. As discussed in Tulsiani \emph{et al.}~\cite{tulsiani2017drc}, using the small set of CADs for both training and test would bias the model toward those samples and therefore is not a recommended benchmark protocol.

\begin{figure*}[t]
\begin{center}
    \includegraphics[width=\linewidth]{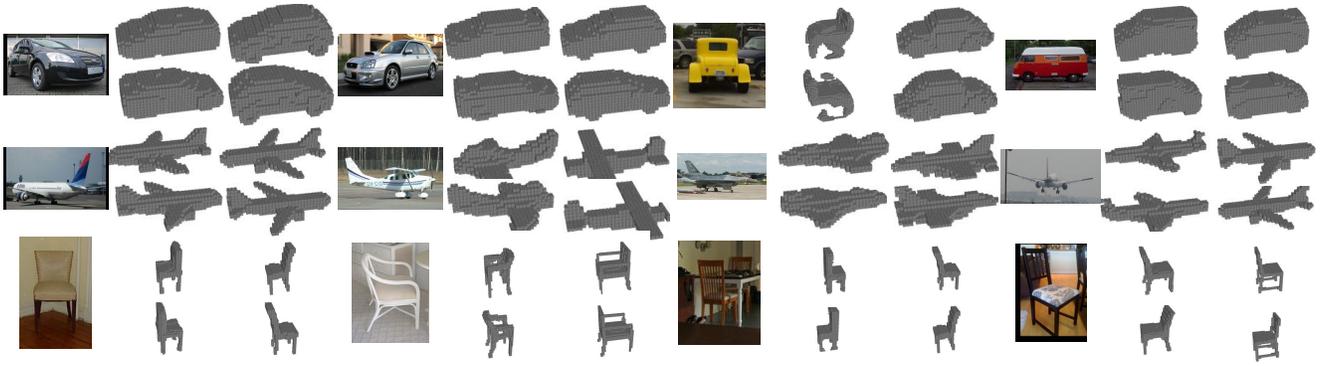}
\end{center}
\caption{3D reconstruction from single image on PASCAL 3D+. For each image, we show the predicted and ground-truth (left) voxel representation (right). We show two different views for each 3D representation.}
\label{fig:pascal-qualitative}
\end{figure*}

We train our model in the categories that are present in both Pascal3D+ and ShapeNet renderings provided by~\cite{choy2016r2n2}:
`aeroplane',`car',`chair',`table' and `tv monitor'. During training, our approach uses ShapeNet rendered images-shape tuples and natural images (we use natural images from ImageNet~\cite{deng09imagenet}). Figure~\ref{fig:pascal-qualitative} shows voxel reconstruction results on different images (and their corresponding ground-truths).

Table~\ref{table:pascal} shows the performance (in terms of CD) of different methods. Following previous work~\cite{tulsiani2017drc,wu2018shapeprior}, we show results in three categories.
Our approach achieves comparable state-of-the-art results with both 3D representations. As before, our reconstruction network, contrary to other methods, does not make use of depths, surface normals, silhouette nor it exploits any form of multiview consistency. Instead, we make use of unlabeled natural images, which are very easy to obtain.
We also note that OGN~\cite{tatarchenko2017octree} uses a much more complex decoder (based on octrees) and considers much higher resolution volumetric occupancy ground-truths during training.

We note that the two novelties of our approach are complementary to previous works and thus could potentially be integrated with those methods for further performance gain.

\begin{table}%[t]
\centering
\begin{tabular}{l|cccc}
% &\multicolumn{4}{c}{Pascal3D+}&
& chair & car & plane & average\\
\shline
3D-R2N2\cite{choy2016r2n2} & 0.238 & 0.305 & 0.305 & 0.284\\
DRC~\cite{tulsiani2017drc} & 0.158 & 0.099 & 0.112 & 0.122\\
OGN~\cite{tatarchenko2017octree} & - & 0.087 & - & -\\
ShapeHD~\cite{wu2018shapeprior} & 0.137 & 0.129 & 0.094 & 0.119\\
\hline
DAREC-vox & 0.135 & 0.101 & 0.108 & 0.115\\
DAREC-pc  & 0.140 & 0.100 & 0.112 & 0.117\\
\end{tabular}
\caption{Single-view 3D reconstruction results on Pascal3D+. We show results on CD metrics. DRC and ShapeHD use depth/normals/silhouettes as extra information during training. OGN considers a much stronger decoder and much higher voxel resolution. Our approach only considers (easily available) natural images.}
\label{table:pascal}
\end{table}

%%%%%%%%%%%%%%%%%%%%%%%%%%%%%%%%%%
% Analyzing the Loss
%%%%%%%%%%%%%%%%%%%%%%%%%%%%%%%%%%
\subsection{Analyzing the Loss}
Here, we perform an ablation study to see how our method performs with respect to different loss terms.
Table~\ref{table:ablation} shows results of our method on Pix3D chair dataset when considering:
% Table~\ref{table:ablation} shows results of our method on Pix3D chair dataset, with models trained on voxel and point clouds representations, when considering:
(i) only the reconstruction loss ($\mathcal{L}_{rec}$), (ii) the reconstruction and the shape prior losses ($\mathcal{L}_{rec}$ and $\mathcal{L}_{shape}$), (iii) the reconstruction and the image domain-confusion losses ($\mathcal{L}_{rec}$ and  $\mathcal{L}_{img}$) and (iv) the full loss.
% In all cases, the network used at inference time has the same architecture (since we do not use the discriminators at test time).
In all those cases, the models have same capacity at inference time (all of them consists of same encoder/decoder architecture).

The first row ($\mathcal{L}_{rec}$ only) ignores the adversarial losses and is trained only with synthetic images. Our method is able to improve the performance by a large margin with both 3D representations (\eg, from .220 to .140 with voxel and .148 to .112 with point cloud).

We first observe that, for both datasets, each loss term has a positive impact on the the final reconstruction result.
The shape prior loss alone is not sufficient to significantly improve the performance. However, the domain confusion loss alone already provides a substantial boost in performance. Finally, the model achieves its best performance when combining both constraints at the same time. These results therefore confirm that each of the proposed loss terms is critical in obtaining the final performance.

\begin{table}[t!]
\centering
\begin{tabular}{ccc|cc}
\multicolumn{3}{c}{}& \multicolumn{2}{c}{Pix3D}\\
$\mathcal{L}_{rec}$ & $\mathcal{L}_{img}$ & $\mathcal{L}_{shape}$ & voxel & point cloud \\
\shline
\checkmark &  &                      & .220 & .148 \\
\checkmark &  & \checkmark           & .196 & .140 \\
\checkmark & \checkmark &            & .156 & .129 \\
\checkmark & \checkmark & \checkmark & .140 & .112 \\
\end{tabular}
% }
\caption{The performance of our model, considering different loss terms, measured with CD on Pix3D chair datasets. We note the importance of each loss component on both metrics, although the shape prior loss alone does not give considerable improvement.}
\label{table:ablation}
\end{table}

%%%%%%%%%%%%%%%%%%%%%%%%%%%%%%%%%%
% Analyzing Learned Representations
%%%%%%%%%%%%%%%%%%%%%%%%%%%%%%%%%%
\subsection{Analyzing the Learned Representations}
\begin{figure}[!t]\centering
\vspace{-0.15cm}
\plottsne{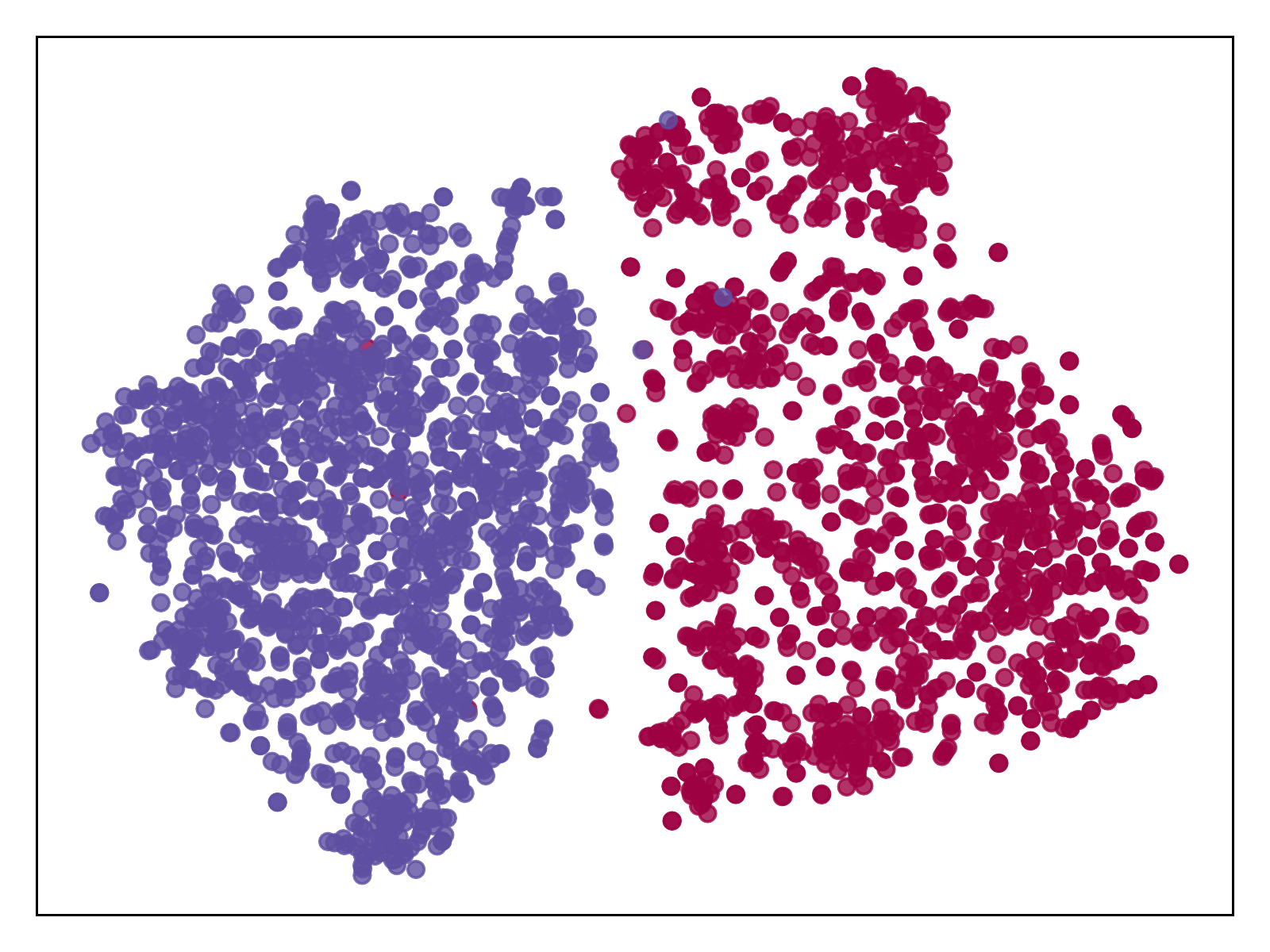}{ }%
\plottsne{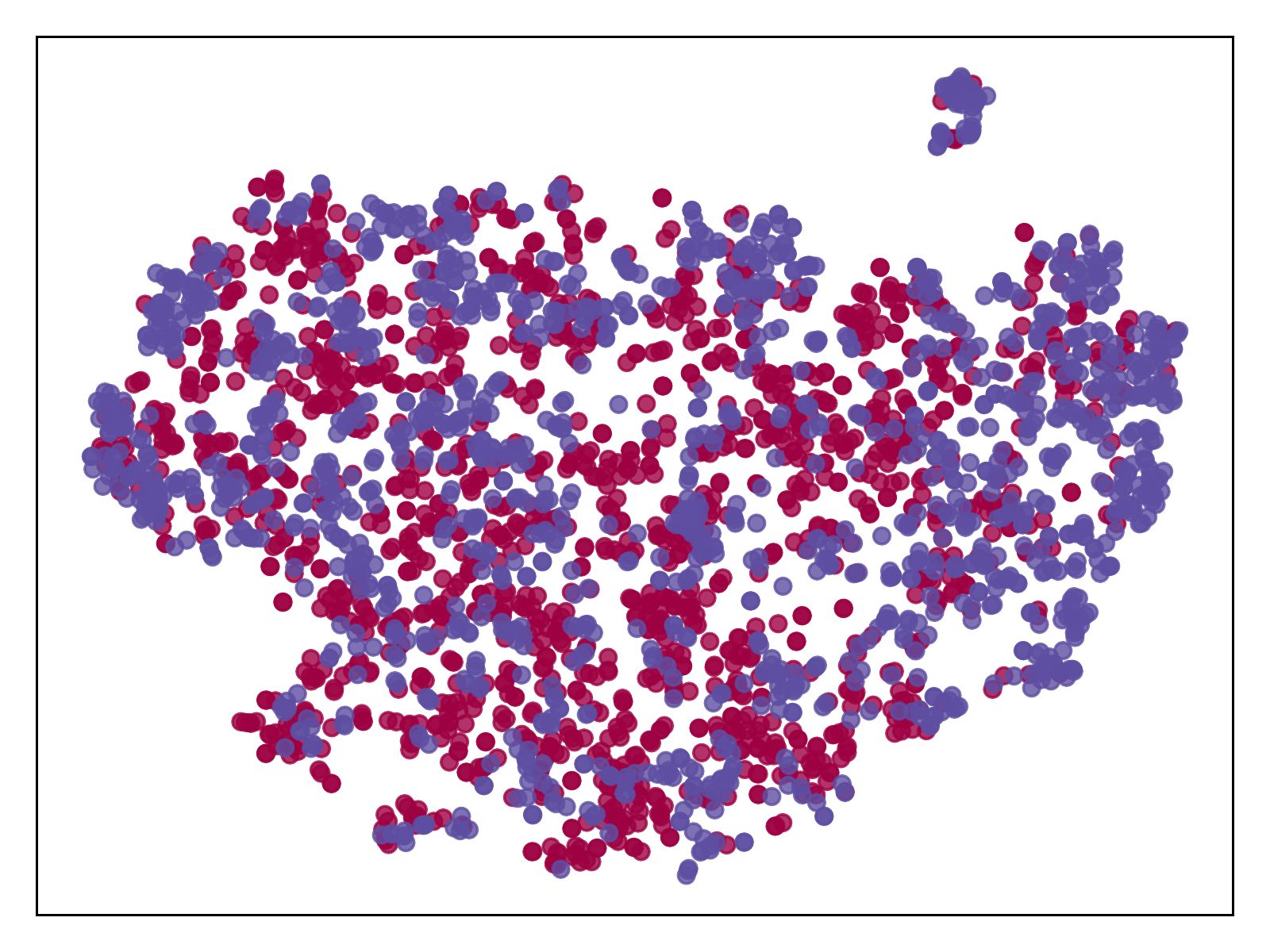}{ }%
\\
\plottsne{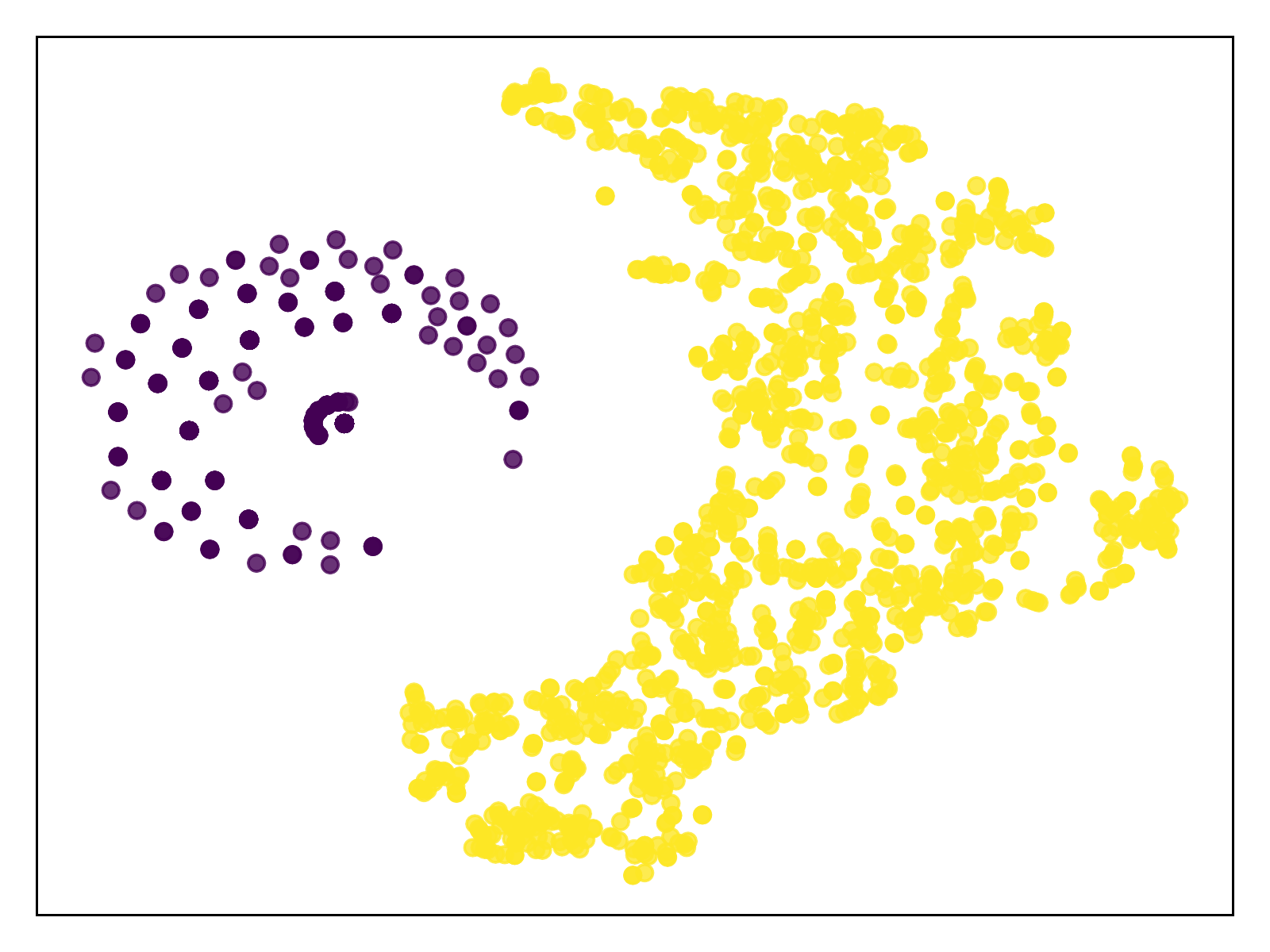}{ }%
\plottsne{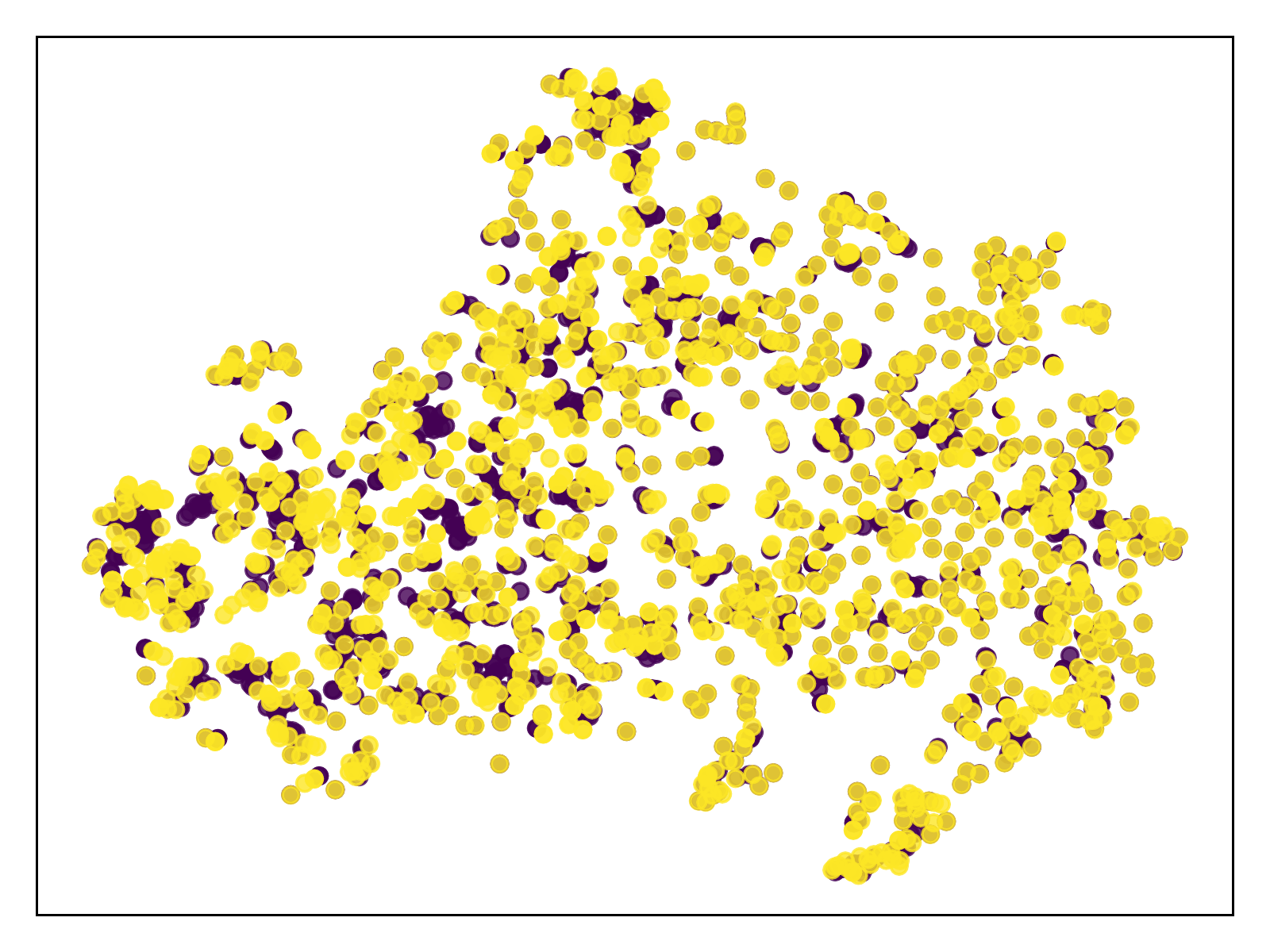}{ }%
\\
\vspace{-0.15cm}
\caption{t-SNE visualization on Pix3D. (a-b) Rendered and natural image embeddings before and after domain confusion. (c-d) 2d rendered embedding and points from manifold before and after training.}
\label{fig:tsne-pix3d}
\vspace{-0.15cm}
\end{figure}

\paragraph{Feature visualization.}
We use t-SNE~\cite{vanDerMaaten2008tsne} to visualize feature representations from different domains and at different adaptation stages (we use the DAREC-vox model on Pix3D). Figure~\ref{fig:tsne-pix3d}(a-b) shows t-SNE features from synthetic (blue) and real (red) images before and after adaptation, respectively. Figure~\ref{fig:tsne-pix3d}(c-d) shows embeddings (before and after training, respectively) of 2D rendered images (blue) and points from the learned shape manifold, \ie, latent representations from the shape autoencoder (yellow).

In both cases, we can see that features become much more domain-invariant after training, as desired. During our experiments, we indeed observed a strong correspondence between reconstruction performance (on natural images) and the overlap between the different feature distributions.

\begin{figure}[!t]
\begin{center}
    \includegraphics[width=\linewidth]{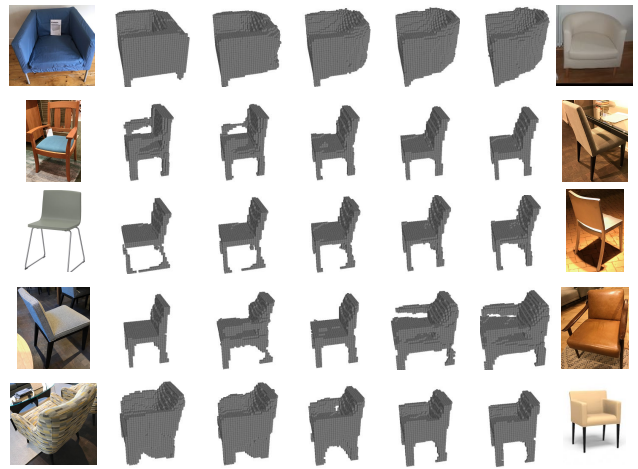}
\end{center}
\caption{Shape interpolation from natural images.}
\label{fig:interpolate}
\end{figure}

\paragraph{Shape interpolation. }
In Figure~\ref{fig:interpolate}, we show results of interpolating between two natural images of different shapes. We first transform each image into its latent representation. Then, we walk through the shape manifold and reconstruct the shape at different interpolated representations. We show qualitatively that the learned shape manifold gives smooth transition between the two object shapes.

\begin{figure}[!t]
\begin{center}
    \includegraphics[width=.8\linewidth]{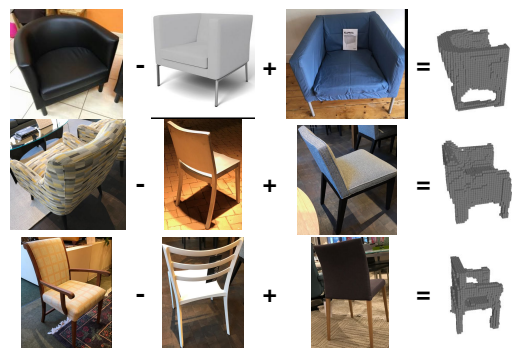}
\end{center}
\caption{Shape arithmetic from natural images. The top row show that `curviness' vector can be added to other chairs. The other rows show that `arm' vector can be added to other chairs.}

\label{fig:arithmetics}
\end{figure}

\vspace{-2mm}
\paragraph{Shape arithmetic. }
Another way to probe the learned representations is to show arithmetic on the latent space. Previous work~\cite{choy2016r2n2,wu2016gan3d,yang2018posesup} showed they are able to learn a semantic manifold of shapes in its latent space and arithmetic is done in samples from this space.
In Figure~\ref{fig:arithmetics}, we perform shape arithmetic on different natural images. We first map them to the learned shape manifold (where arithmetical operations are done), then we reconstruct its shape. We observe that the representation after the arithmetic operations are still reasonable to reconstruct a realistic shape.

%%%%%%%%%%%%%%%%%%%%%%%%%%%%%%%%%%%%%%%%%%%%%%%%%%%%%%%%%%%%%%%%%%%%%%%%%%%%%%%
% Conclusion
%%%%%%%%%%%%%%%%%%%%%%%%%%%%%%%%%%%%%%%%%%%%%%%%%%%%%%%%%%%%%%%%%%%%%%%%%%%%%%%
\section{Conclusion}\label{sec:conclusion}
In this paper, we presented a framework for improved 3D reconstruction from  single-view natural image. Our method leverages adversarial training and shape priors in two different ways. First it imposes learned features to be domain-invariant to help with the problem of domain adaptation. Second, we force the learned representations to lie in a rich shape prior manifold, imposing the reconstructions to be realistic. We show our method is able to improve the performance when considering different 3D representations. By using only RGB signal and with a much simpler network architecture, our model achieves competitive performance with the state of the art.

{\small
\bibliographystyle{ieee_fullname}
\bibliography{paper}

\begin{thebibliography}{10}\itemsep=-1pt

\bibitem{bendavid2010da}
Shai Ben-David, John Blitzer, Koby Crammer, Alex Kulesza, Fernando Pereira, and
  Jennifer~Wortman Vaughan.
\newblock A theory of learning from different domains.
\newblock {\em Machine Learning}, 2010.

\bibitem{bendavid2007da}
Shai Ben-David, John Blitzer, Koby Crammer, and Fernando Pereira.
\newblock Analysis of representations for domain adaptation.
\newblock In {\em NIPS}, 2007.

\bibitem{blanz99morphable}
Volker Blanz and Thomas Vetter.
\newblock A morphable model for the synthesis of 3d faces.
\newblock In {\em SIGGRAPH}, 1999.

\bibitem{bonetV99}
Jeremy S.~De Bonet and Paul~A. Viola.
\newblock Roxels: Responsibility weighted 3d volume reconstruction.
\newblock In {\em ICCV}, 1999.

\bibitem{bridle90softmax}
John Bridle.
\newblock Probabilistic interpretation of feedforward classification network
  outputs, with relationships to statistical pattern recognition.
\newblock {\em Neurocomputing: Algorithms, Architectures and Applications},
  1990.

\bibitem{broadhurst2001APF}
Adrian Broadhurst, Tom Drummond, and Roberto Cipolla.
\newblock A probabilistic framework for space carving.
\newblock In {\em ICCV}, 2001.

\bibitem{chang2015shapenet}
Angel~X. Chang, Thomas~A. Funkhouser, Leonidas~J. Guibas, Pat Hanrahan, Qi-Xing
  Huang, Zimo Li, Silvio Savarese, Manolis Savva, Shuran Song, Hao Su,
  Jianxiong Xiao, Li Yi, and Fisher Yu.
\newblock Shapenet: An information-rich 3d model repository.
\newblock In {\em CoRR}, 2015.

\bibitem{choy2016r2n2}
Christopher~Bongsoo Choy, Danfei Xu, JunYoung Gwak, Kevin Chen, and Silvio
  Savarese.
\newblock 3d-r2n2: {A} unified approach for single and multi-view 3d object
  reconstruction.
\newblock In {\em ECCV}, 2016.

\bibitem{csurka17uda}
Gabriela Csurka.
\newblock A comprehensive survey on domain adaptation for visual applications.
\newblock In {\em Domain Adaptation in Computer Vision Applications}, Advances
  in Computer Vision and Pattern Recognition. Springer, 2017.

\bibitem{deng09imagenet}
Jia Deng, Wei Dong, Richard Socher, Li-Jia Li, Kai Li, and Li Fei-Fei.
\newblock Imagenet: A large-scale hierarchical image database.
\newblock In {\em CVPR}, 2009.

\bibitem{eslami18gqn}
S.~M.~Ali Eslami, Danilo~Jimenez Rezende, Fr{\'e}d{\'e}ric Besse, Fabio Viola,
  Ari~S. Morcos, Marta Garnelo, Avraham Ruderman, Andrei~A. Rusu, Ivo
  Danihelka, Karol Gregor, David~P. Reichert, Lars Buesing, Theophane Weber,
  Oriol Vinyals, Dan Rosenbaum, Neil~C. Rabinowitz, Helen King, Chloe Hillier,
  Matthew~M Botvinick, Daan Wierstra, Koray Kavukcuoglu, and Demis Hassabis.
\newblock Neural scene representation and rendering.
\newblock {\em Science}, 2018.

\bibitem{everingham2015pascal}
Mark Everingham, S.~M. Eslami, Luc Gool, Christopher~K. Williams, John Winn,
  and Andrew Zisserman.
\newblock The pascal visual object classes challenge: A retrospective.
\newblock {\em IJCV}, 2015.

\bibitem{fan2017points}
Haoqiang Fan, Hao Su, and Leonidas~J. Guibas.
\newblock A point set generation network for 3d object reconstruction from a
  single image.
\newblock In {\em CVPR}, 2017.

\bibitem{furukawa2015multiview}
Yasutaka Furukawa and Carlos Hernández.
\newblock Multi-view stereo: A tutorial.
\newblock {\em Foundations and Trends in Computer Graphics and Vision}, 2015.

\bibitem{ganin16revgrad}
Yaroslav Ganin, Evgeniya Ustinova, Hana Ajakan, Pascal Germain, Hugo
  Larochelle, Fran{\c{c}}ois Laviolette, Mario Marchand, and Victor~S.
  Lempitsky.
\newblock Domain-adversarial training of neural networks.
\newblock {\em JMLR}, 2016.

\bibitem{girdhar16tnet}
Rohit Girdhar, David~F. Fouhey, Mikel Rodriguez, and Abhinav Gupta.
\newblock Learning a predictable and generative vector representation for
  objects.
\newblock In {\em ECCV}, 2016.

\bibitem{goodfellow16dl}
Ian Goodfellow, Yoshua Bengio, and Aaron Courville.
\newblock {\em Deep Learning}.
\newblock The MIT Press, 2016.

\bibitem{goodfellow2014gan}
Ian Goodfellow, Jean Pouget-Abadie, Mehdi Mirza, Bing Xu, David Warde-Farley,
  Sherjil Ozair, Aaron Courville, and Yoshua Bengio.
\newblock Generative adversarial nets.
\newblock In {\em NIPS}, 2014.

\bibitem{groueix2018atlas}
Thibault Groueix, Matthew Fisher, Vladimir~G. Kim, Bryan~C. Russell, and
  Mathieu Aubry.
\newblock Atlasnet: {A} papier-m{\^{a}}ch{\'{e}} approach to learning 3d
  surface generation.
\newblock 2018.

\bibitem{gwak17weakly}
JunYoung Gwak, Christopher~B. Choy, Animesh Garg, Manmohan Chandraker, and
  Silvio Savarese.
\newblock Weakly supervised generative adversarial networks for 3d
  reconstruction.
\newblock In {\em 3DV}, 2017.

\bibitem{he16resnet}
Kaiming He, Xiangyu Zhang, Shaoqing Ren, and Jian Sun.
\newblock Deep residual learning for image recognition.
\newblock In {\em CVPR}, 2016.

\bibitem{rezende2016structure3d}
Danilo Jimenez~Rezende, S.~M.~Ali Eslami, Shakir Mohamed, Peter Battaglia, Max
  Jaderberg, and Nicolas Heess.
\newblock Unsupervised learning of 3d structure from images.
\newblock In {\em NIPS}, 2016.

\bibitem{kanazawaT18collections}
Angjoo Kanazawa, Shubham Tulsiani, Alexei~A. Efros, and Jitendra Malik.
\newblock Learning category-specific mesh reconstruction from image
  collections.
\newblock In {\em ECCV}, 2018.

\bibitem{kar2015csrec}
Abhishek Kar, Shubham Tulsiani, João Carreira, and Jitendra Malik.
\newblock Category-specific object reconstruction from a single image.
\newblock In {\em CVPR}, 2015.

\bibitem{kingma14adam}
Diederik~P. Kingma and Jimmy Ba.
\newblock Adam: A method for stochastic optimization.
\newblock {\em ICLR}, 2014.

\bibitem{kourtzi11neural}
Zoe Kourtzi and Charles~E Connor.
\newblock Neural representations for object perception: structure, category,
  and adaptive coding.
\newblock {\em Annual review of neuroscience}, 2011.

\bibitem{kundu18rcnn3d}
Abhijit Kundu, Yin Li, and James~M. Rehg.
\newblock 3d-rcnn: Instance-level 3d object reconstruction via
  render-and-compare.
\newblock In {\em CVPR}, 2018.

\bibitem{laurentini1994hull}
Aldo Laurentini.
\newblock The visual hull concept for silhouette-based image understanding.
\newblock {\em PAMI}, 1994.

\bibitem{lecun98cnn}
Yann Lecun, Léon Bottou, Yoshua Bengio, and Patrick Haffner.
\newblock Gradient-based learning applied to document recognition.
\newblock In {\em Proceedings of the IEEE}, 1998.

\bibitem{li15joint}
Yangyan Li, Hao Su, Charles~Ruizhongtai Qi, Noa Fish, Daniel Cohen-Or, and
  Leonidas~J. Guibas.
\newblock Joint embeddings of shapes and images via cnn image purification.
\newblock {\em ACM Trans. Graph.}, 2015.

\bibitem{makhzani2016aae}
Alireza Makhzani, Jonathon Shlens, Navdeep Jaitly, and Ian Goodfellow.
\newblock Adversarial autoencoders.
\newblock In {\em ICLR}, 2016.

\bibitem{nair10relu}
Vinod Nair and Geoffrey~E. Hinton.
\newblock Rectified linear units improve restricted boltzmann machines.
\newblock In {\em ICML}, 2010.

\bibitem{novotny2017look}
David Novotn{\'{y}}, Diane Larlus, and Andrea Vedaldi.
\newblock Learning 3d object categories by looking around them.
\newblock In {\em ICCV}, 2017.

\bibitem{pinheiro18simnet}
Pedro~O. Pinheiro.
\newblock Unsupervised domain adaptation with similarity learning.
\newblock In {\em CVPR}, 2018.

\bibitem{qi2017pointnet}
Charles~R Qi, Hao Su, Kaichun Mo, and Leonidas~J Guibas.
\newblock Pointnet: Deep learning on point sets for 3d classification and
  segmentation.
\newblock In {\em CVPR}, 2017.

\bibitem{saito18advdrop}
Kuniaki Saito, Yoshitaka Ushiku, Tatsuya Harada, and Kate Saenko.
\newblock Adversarial dropout regularization.
\newblock In {\em ICLR}, 2018.

\bibitem{sait18maxdisc}
Kuniaki Saito, Kohei Watanabe, Yoshitaka Ushiku, and Tatsuya Harada.
\newblock Maximum classifier discrepancy for unsupervised domain adaptation.
\newblock In {\em CVPR}, 2018.

\bibitem{saito18openset}
Kuniaki Saito, Shohei Yamamoto, Yoshitaka Ushiku, and Tatsuya Harada.
\newblock Open set domain adaptation by backpropagation.
\newblock In {\em ECCV}, 2018.

\bibitem{seitz2006}
Steven~M. Seitz, Brian Curless, James Diebel, Daniel Scharstein, and Richard
  Szeliski.
\newblock A comparison and evaluation of multi-view stereo reconstruction
  algorithms.
\newblock In {\em CVPR}, 2006.

\bibitem{sun2018pix3d}
Xingyuan Sun, Jiajun Wu, Xiuming Zhang, Zhoutong Zhang, Chengkai Zhang, Tianfan
  Xue, Joshua~B. Tenenbaum, and William~T. Freeman.
\newblock Pix3d: Dataset and methods for single-image 3d shape modeling.
\newblock In {\em CVPR}, 2018.

\bibitem{tatarchenko2016multiview}
Maxim Tatarchenko, Alexey Dosovitskiy, and Thomas Brox.
\newblock Multi-view 3d models from single images with a convolutional network.
\newblock In {\em ECCV}, 2016.

\bibitem{tatarchenko2017octree}
Maxim Tatarchenko, Alexey Dosovitskiy, and Thomas Brox.
\newblock Octree generating networks: Efficient convolutional architectures for
  high-resolution 3d outputs.
\newblock In {\em ICCV}, 2017.

\bibitem{torralba11cvpr}
Antonio Torralba and Alexei~A. Efros.
\newblock Unbiased look at dataset bias.
\newblock In {\em CVPR}, 2011.

\bibitem{tulsiani2017drc}
Shubham Tulsiani, Tinghui Zhou, Alexei~A. Efros, and Jitendra Malik.
\newblock Multi-view supervision for single-view reconstruction via
  differentiable ray consistency.
\newblock In {\em {CVPR}}, 2017.

\bibitem{vanDerMaaten2008tsne}
Laurens van~der Maaten and Geoffrey Hinton.
\newblock Visualizing data using {t-SNE}.
\newblock {\em JMLR}, 2008.

\bibitem{vicente14recvoc}
Sara Vicente, Jo\~{a}o Carreira, Lourdes Agapito, and Jorge Batista.
\newblock Reconstructing pascal voc.
\newblock In {\em CVPR}, 2014.

\bibitem{wu2017marrnet}
Jiajun Wu, Yifan Wang, Tianfan Xue, Xingyuan Sun, Bill Freeman, and Josh
  Tenenbaum.
\newblock Marrnet: 3d shape reconstruction via 2.5d sketches.
\newblock In {\em {NIPS}}, 2017.

\bibitem{wu2016gan3d}
Jiajun Wu, Chengkai Zhang, Tianfan Xue, Bill Freeman, and Josh Tenenbaum.
\newblock Learning a probabilistic latent space of object shapes via 3d
  generative-adversarial modeling.
\newblock In {\em NIPS}, 2016.

\bibitem{wu2018shapeprior}
Jiajun Wu, Chengkai Zhang, Xiuming Zhang, Zhoutong Zhang, William~T. Freeman,
  and Joshua~B. Tenenbaum.
\newblock Learning shape priors for single-view 3d completion and
  reconstruction.
\newblock In {\em ECCV}, 2018.

\bibitem{wu2015objectnet}
Zhirong Wu, Shuran Song, Aditya Khosla, Fisher Yu, Linguang Zhang, Xiaoou Tang,
  and Jianxiong Xiao.
\newblock 3d shapenets: A deep representation for volumetric shapes.
\newblock In {\em CVPR}, 2015.

\bibitem{xiang2016objectnet3d}
Yu Xiang, Wonhui Kim, Wei Chen, Jingwei Ji, Christopher Choy, Hao Su, Roozbeh
  Mottaghi, Leonidas Guibas, and Silvio Savarese.
\newblock Objectnet3d: A large scale database for 3d object recognition.
\newblock In {\em ECCV}, 2016.

\bibitem{xiang14pascal3d}
Yu Xiang, Roozbeh Mottaghi, and Silvio Savarese.
\newblock Beyond pascal: A benchmark for 3d object detection in the wild.
\newblock In {\em WACV}, 2014.

\bibitem{yukako08shape}
Yukako Yamane, Eric~T Carlson, Katherine~C Bowman, Zhihong Wang, and Charles~E
  Connor.
\newblock A neural code for three-dimensional object shape in macaque
  inferotemporal cortex.
\newblock {\em Nature Neuroscience}, 2008.

\bibitem{xinchen2016perspective}
Xinchen Yan, Jimei Yang, Ersin Yumer, Yijie Guo, and Honglak Lee.
\newblock Perspective transformer nets: Learning single-view 3d object
  reconstruction without 3d supervision.
\newblock In {\em NIPS}, 2016.

\bibitem{yang2018posesup}
Guandao Yang, Yin Cui, Serge Belongie, and Bharath Hariharan.
\newblock Learning single-view 3d reconstruction with limited pose supervision.
\newblock In {\em ECCV}, 2018.

\bibitem{zhang18unseen}
Xiuming Zhang, Zhoutong Zhang, Chengkai Zhang, Josh Tenenbaum, Bill Freeman,
  and Jiajun Wu.
\newblock Learning to reconstruct shapes from unseen classes.
\newblock In {\em NeurIPS}. 2018.

\end{thebibliography}
}

\end{document}